\documentclass[review]{elsarticle}

\usepackage{amsmath, amssymb}
\usepackage{hyperref}

\usepackage{algorithm}
\usepackage{algpseudocodex}

\usepackage{booktabs}
\usepackage{graphicx}
\usepackage{array}

\usepackage{import}

\biboptions{sort&compress} 

\usepackage{caption}
\usepackage{subcaption}

\newcommand{\etal}{\textit{et al.}} 

\usepackage[top=25mm,inner=20mm,outer=20mm,bottom=25mm]{geometry}

\begin{document}

\begin{frontmatter}



\title{A spectral regularisation framework for latent variable models designed for single channel applications}


\author[mymainaddress]{Ryan Balshaw} 

\author[mymainaddress]{P. Stephan Heyns}

\author[mymainaddress]{Daniel N. Wilke\corref{mycorrespondingauthor}}
\cortext[mycorrespondingauthor]{Corresponding author}
\ead{nico.wilke@up.ac.za}

\author[mymainaddress]{Stephan Schmidt}

\address[mymainaddress]{Centre for Asset Integrity Management, Department of Mechanical and Aeronautical Engineering, University of Pretoria, Pretoria, South Africa}

\begin{abstract}
Latent variable models (LVMs) are commonly used to capture the underlying dependencies, patterns, and hidden structure in observed data. Source duplication is a by-product of the data hankelisation pre-processing step common to single channel LVM applications, which hinders practical LVM utilisation. In this article, a Python package titled \texttt{spectrally-regularised-LVMs} is presented. The proposed package addresses the source duplication issue via the addition of a novel spectral regularisation term. This package provides a framework for spectral regularisation in single channel LVM applications, thereby making it easier to investigate and utilise LVMs with spectral regularisation. This is achieved via the use of symbolic or explicit representations of potential LVM objective functions which are incorporated into a framework that uses spectral regularisation during the LVM parameter estimation process. The objective of this package is to provide a consistent linear LVM optimisation framework which incorporates spectral regularisation and caters to single channel time-series applications.

\end{abstract}


\end{frontmatter}


%
%

\section{Introduction}\label{sec:introduction}
\begin{figure*}[htb!]
	\centering
	\includegraphics[width=\textwidth]{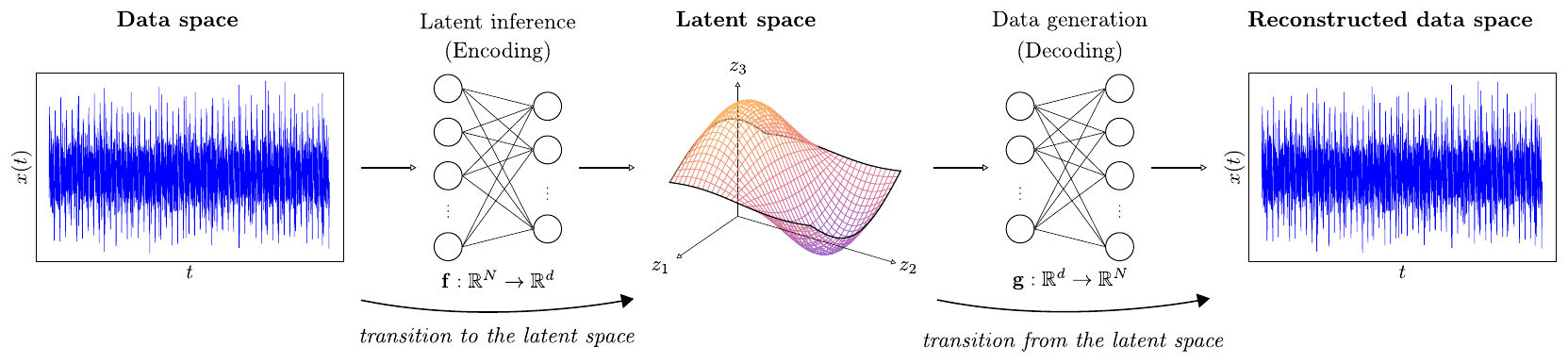}
	\caption{A pictorial representation of the LVM framework applied to single channel time-series data. Note that the dimensionality of the latent space is presented in $\mathbb{R}^3$ and no data pre-processing steps are considered between the data space and the latent space. Data pre-processing steps are discussed in Section \ref{subsec:pre-processing}.}
	\label{fig:LVM_example}
\end{figure*}

Latent variable models (LVMs) represent a statistical methodology which aims to extract hidden sources of information, referred to as the latent variables $\mathbf{z} \sim p(\mathbf{z}),\, \mathbf{z} \in \mathbb{R}^{d}$, from an observed set of random variables $\mathbf{x} \sim p(\mathbf{x}),\, \mathbf{x} \in \mathbb{R}^{D}$. The overarching goal of the LVM framework, as demonstrated pictorially via Figure \ref{fig:LVM_example}, is to \emph{i)} construct a generative model to generate observed samples, and \emph{ii)} infer the latent variables, commonly referred to as the latent sources, from the observed data~\cite{Blei2014, Blei2017}. In the linear LVM formulation, a linear encoding and decoding step is used to obtain samples from the posterior distribution $p(\mathbf{z} \vert \mathbf{x})$ and the generative distribution $p(\mathbf{x} \vert \mathbf{z})$. 

In common LVM frameworks, e.g., principal component analysis (PCA)~\cite{PearsonK.1901, Hotelling1933, Bishop2006, Tipping1999}, a Gaussian assumption is made for the prior $p(\mathbf{z})$, the generative distribution $p(\mathbf{x}\vert\mathbf{z})$ and posterior distribution $p(\mathbf{z} \vert \mathbf{x})$, and the latent sources are driven to capture components that best describe the dominant sources in the observed data via minimum mean-square error compression or variance maximisation~\cite{Bishop2006}. In this way, these methodologies are sample generation focused and the latent sources capture the dominant variance in the observed data. The PCA framework deals with secord-order statistics and results in uncorrelated latent sources, i.e. the covariance for two sources $z_i$ and $z_j$, represented by $\text{cov}\left[z_i, z_j\right]$, for $i \neq j$ is zero and non-zero otherwise. The independent component analysis (ICA) formulation manifests as a LVM with latent sources that are assumed as non-Gaussian and mutually independent~\cite{Jutten1991, Comon1994, Bell1995, Pham1997}. In ICA frameworks, the assumptions of statistical independence and non-Gaussianity are more rigorous and enforce that the latent sources should be maximally informative in a non-Gaussian setting~\cite{Hyvarinen2000, Hyvarinen2001a}. The statistical independence assumption implies that the latent distribution factorises, i.e. $p(\mathbf{z}) = \prod_{i=1}^{d} p_{i}(z_i)$, where $\mathbf{z} \in \mathbb{R}^{d}$. In the assumed-Gaussian setting, it is possible to show that the PCA linear transformation can produce a statistically independent latent distribution~\cite{Bishop2006}.

In applications when a single channel time-series signal is of interest and available, certain processing steps, e.g., data hankelisation, must be taken to fully enable the LVM framework~\cite{Balshaw2022}. In this setting, as noted in Hyv\"{a}rinen~\cite{Hyvarinen2001a}, the linear LVM framework is strongly related to the blind-deconvolution problem~\cite{Haykin2014}, and the linear model formulation results in an underdetermined source situation, i.e., the sources extracted by the model may contain duplicate information~\cite{Hyvarinen2001a}. To fully recover the latent sources, Davies and James~\cite{Davies2007} demonstrated that the original sources in the observed data must have disjoint spectral support. Additionally, if no modifications are made to the standard LVM objective functions, clustering algorithms must be used to combine extracted sources that contain duplicate information~\cite{Davies2007}. Source duplication is an existing limitation that hinders the practicality of certain LVM objective functions in single channel time-series applications as the full rank solution, i.e. $d = D$, is required to avoid missing any interesting sources. Additionally, the model estimation step may be computationally extensive if $d << D$ as much compute is wasted on duplicate information. Hence, we seek to bypass this limitation to improve the practicality of linear LVMs. In this work, a framework is proposed which offers ease of objective function implementation and a spectral regularisation term is proposed and implemented to enforce the disjoint spectral support requirement. 

The proposed framework caters to spectral regularisation for linear LVM formulations and the \texttt{spectrally-\\regularised-LVMs} package makes spectral regularisation readily accessible for first order and second order optimisers. The package provides simplicity, ease of use, and efficiency to the single channel LVM framework, whereby the burden of repetitive pre-processing and optimisation framework implementation tasks are no longer placed on the user. The proposed framework operates in a deflation-based optimisation setting, i.e. each latent source is solved for one-by-one. Additionally, the methodology is objective function agnostic, which allows users to implement their own cost functions and use the proposed framework. Users can implement their objective functions directly using the symbolic Python package SymPy~\cite{Sympy2017} to automatically obtain the gradient and Hessian of the model objective function, or using standard linear algebra operations enabled though the NumPy library~\cite{Harris2020}. Additionally, a finite-difference scheme is provided should users wish to obtain approximations of the gradient and Hessian of the model objective function. This provides generality to the \texttt{spectrally-regularised-LVMs} package and creates flexibility for future development and package utilisation.

\subsection{Related work}
In the literature, spectral regularisation typically refers to a regularisation technique for learning-based problems which utilises the spectra of an operator or matrix to control the properties of some model~\cite{Rosasco2006}, e.g. Tikhonov regularisation~\cite{tikhonov1977}. For matrices, the spectra refers to its eigenvalue set~\cite{golub2013}. Gerfon \etal~\cite{Gerfon2008a} provide a succinct overview of spectral regularisation techniques for supervised learning tasks. Miyato \etal~\cite{Miyato2018} use the spectral properties of deep generative models to improve model performance. Specifically,~\cite{Miyato2018} use the spectral norm of weight matrices to control the Lipschitz continuity of the discriminator network function for generative adversarial networks to improve model trainability. Xie \etal~\cite{Xie2019a} use a spectral loss based off cosine similarity to encourage reconstruction consistency for anomaly detection using adversarial auto-encoders. In this work, spectral regularisation refers to the regularisation of the spectra of the LVM component vectors, where the spectra is given by the Fourier transform, to ensure that it does not capture the same signal information as other component vectors. In this way, spectral orthogonality within the LVM components is encouraged and the signal information captured by the LVM is driven to consist of unique components.

\section{Model formulation}
In this section the pre-processing steps for single channel time-series are detailed, the preliminary formulation of linear LVMs is presented, and the proposed spectral regularisation term is defined.

\subsection{Time-series pre-processing}\label{subsec:pre-processing}
For single channel time-series data, we obtain a signal $x[n]$, where $n \subset \mathbb{Z}^{+}= \{1, \cdots, L\}$ and $L$ is the length of the signal, that must be pre-processed to fit into the standard LVM framework. This can be achieved through data hankelisation to obtain a matrix $\mathbf{X}\in\mathbb{R}^{L_H \times L_w}$
\begin{equation}\label{eq:Hankel_Mat}
	\mathbf{X} = 
	\begin{bmatrix}
		x[1] & \cdots & x[L_w] \\
		x[L_{sft}] & \cdots & x[L_{sft} + L_w] \\
		x[2 \cdot L_{sft}] & \cdots & x[2 \cdot L_{sft} + L_w] \\
		\vdots & \ddots  & \vdots \\
		x[L_{sft} \cdot (L_H - 1)] & \cdots & x[L_{sft} \cdot (L_H - 1) + L_w] \\
	\end{bmatrix}
\end{equation}
where $L_w$ is the window length, $L_{sft}$ is the shift parameter, and $L_H = \lfloor\frac{L - L_w}{L_{sft}}\rfloor + 1$ represents the number of rows in $\mathbf{X}$. Note that the shift parameter is typically set to $L_{sft} = 1$ to ensure that the LVM component vectors act as a set of finite impulse response (FIR) filters~\cite{Davies2007}.

\subsection{Linear latent variable models}
The basic linear LVM generative model may be expressed for some observed data $\mathbf{x} \in \mathbb{R}^{D}$ as
\begin{equation}
	\mathbf{x} = \sum_{i=1}^{d} \mathbf{a}_i z_i,
\end{equation}
where $d \leq D$, $z_i$ represents the $i^{th}$ latent source and $\mathbf{a}_i \in \mathbb{R}^{D}$ represents the $i^{th}$ decoding transition vector. The matrix-vector notation of the generative model is
\begin{equation}\label{eq:data-transition}
	\mathbf{x} = \mathbf{A} \mathbf{z},
\end{equation} 
where $\mathbf{A} \in \mathbb{R}^{D \times d} = \left[\mathbf{a}_1, \cdots, \mathbf{a}_d \right]$. In the latent encoding step, a matrix $\mathbf{W} \in \mathbb{R}^{d \times D} = [\mathbf{w}_1, \cdots, \mathbf{w}_d]^T$ can be used to infer the latent vector $\mathbf{z}$ by
\begin{equation}\label{eq:latent-transition}
	\mathbf{z} = \mathbf{W} \mathbf{x},
\end{equation}
where the rows in $\mathbf{W}$ represent the encoding transition vectors. In the LVM model, samples $\mathbf{x} \sim p(\mathbf{x})$ are observed from some unknown distribution $p(\mathbf{x})$, and the goal is to estimate $\mathbf{A}$ or $\mathbf{W}$. Common methodologies prefer to utilise characteristics of the latent sources to estimate the model parameters, e.g., maximum latent variance~\cite{Hotelling1933} or maximum non-Gaussianity through measures such as kurtosis or negentropy~\cite{Hyvarinen2000}. The objective function of interest is henceforward generally denoted by $\mathcal{L}_{model}$.

\subsection{Spectral regularisation}
In this work, we propose using a spectral regularisation term for linear LVMs objective functions to combat source duplication and use a general optimisation framework to estimate the model parameters. The assumed form of the general LVM objective function may be written as
\begin{equation}\label{eq:objective-function}
	\begin{aligned}[b]
		\min_{\mathbf{w}_i} \quad & \mathcal{L}_{model}(\mathbf{w}_i) + \mathcal{L}_{sr}(\mathbf{w}_i) \\
		\text{s.t.} \quad & \mathbf{w}_i^T\mathbf{w}_i = 1,
	\end{aligned}
\end{equation}
where $\mathcal{L}_{model}(\mathbf{w}_i)$ represents the objective function to be minimised, $\mathcal{L}_{sr}(\mathbf{w}_i)$ represents the spectral regulariser that acts as an explicit regularisation term on $\mathcal{L}_{model}(\mathbf{w}_i)$, and the optimisation constraint $\mathbf{w}_i^T\mathbf{w}_i=1$ is used to ensure that the parameter estimation process focuses on the direction of $\mathbf{w}$ and not its magnitude. The spectral regularisation term encourages the $\mathbf{w}_i$ solution to be spectrally unique with respect to the previous solution vectors $\mathbf{w}_j$, where $j < i$. The general objective function can be reformulated into a Lagrangian expression
\begin{equation}\label{eq:Lagrangian-function-text}
	\begin{aligned}[b]
	\mathcal{L}(\mathbf{w}_i, \lambda_{eq}) &= \mathcal{L}_{model}(\mathbf{w}_i) + \mathcal{L}_{sr}(\mathbf{w}_i) \\
	& + \lambda_{eq} \left( \mathbf{w}_i^T \mathbf{w}_i - 1 \right),
	\end{aligned}
\end{equation}
where $\lambda_{eq}$ represents the Lagrange multiplier. The regularisation term proposed in this work enforces spectral orthogonality and uses the squared modulus of the Fourier domain representation of a vector $\mathbf{w}_{i}$, denoted as $\mathbf{b}(\mathbf{w}_{i})$, to minimise the dot product between $\mathbf{b}(\mathbf{w}_{i})$ and the squared Fourier representation for the previously solved $\mathbf{w}_j$ vectors, $j = 1, \cdots, i - 1$. This is given as

\begin{equation}\label{eq:constraint}
	\mathcal{L}_{sr}(\mathbf{w}_i) = \alpha \sum_{j=1}^{i -1} \mathbf{b}(\mathbf{w}_{i})^T \mathbf{b}(\mathbf{w}_{j}), \quad i > 1
\end{equation}
where $\alpha$ represents a penalty enforcement parameter that controls the importance of the regularisation term. Note that the spectral regularisation term is only applied after the estimation of the first projection vector $\mathbf{w}_1$. To remove ambiguity behind the choice of $\alpha$, the sequential unconstrained minimisation technique (SUMT) is used within the model optimisation step to iteratively increase the prominence of the regularisation term during each optimisation iteration~\cite{Snyman2018}. The important terms related to the implemented optimisation algorithm to incorporate the regularisation term in Equation \ref{eq:constraint} into the overall LVM objective are detailed in the \href{https://spectrally-regularised-lvms.readthedocs.io/en/latest/}{\textcolor{blue}{documentation}}. The full derivation is not included here for brevity. The use of the spectral regularisation term in Equation \eqref{eq:constraint} allows the basic LVM model to be effectively applied to single channel time-series data. There is no reliance on carefully selected pre-processing strategies, e.g., the spectrogram, and source duplication is discouraged during the parameter estimation process.  

\section{Software description}
\begin{figure*}
	\centering
	\includegraphics[width=0.6\textwidth]{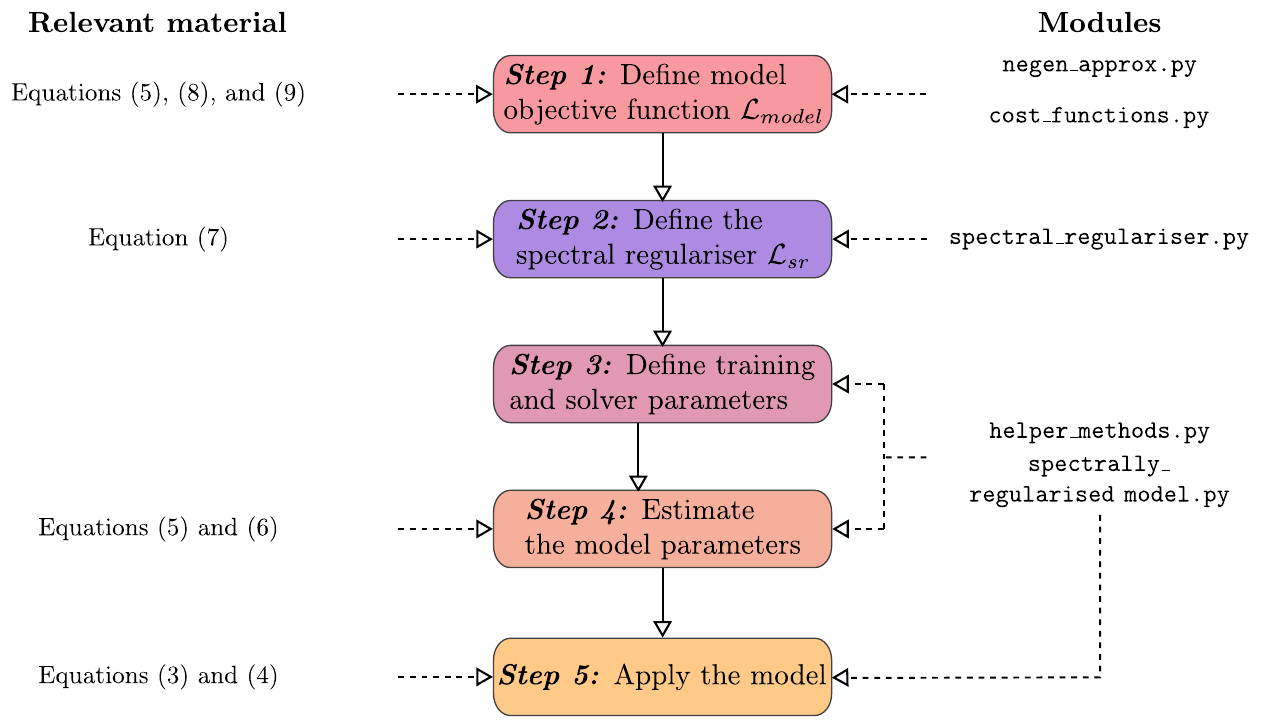}
	\caption{A summary of the important mathematical aspects that provide context to the software implementation and architecture, the associated package sub-modules, and any relevant material that is provided in this document.}
	\label{fig:package_description}
\end{figure*}

In this section, a basic description of the software used in the \texttt{spectrally-regularised-LVMs} Python package is provided. Figure \ref{fig:package_description} provides context to how the relevant mathematical components of the package can be sequentially organised.

\subsection{Introduction to spectrally-regularised-LVMs}\label{subsec:introduction}
The \texttt{spectrally-regularised-LVMs} package provides a framework for linear LVMs with the proposed spectral regularisation term using the Python API. In this framework, the objective function for the LVM can either be implemented symbolically or explicitly as a Python function that uses NumPy objects. Alternatively, the objective function for two common LVMs, PCA and ICA, can be called through the sub-modules present in the package. The choice of a symbolic objective function implementation ensures that users need not spend effort on deriving the gradient and Hessian of their objective function, as this can be done symbolically. For the explicit implementation, users can either provide the necessary first and second order information, or use a finite difference approximation scheme to estimate the gradient and Hessian. As detailed in Figure \ref{fig:package_description}, there are five steps that are required to formalise the proposed LVM framework in single channel applications. These five steps are combined within the package to create a framework from which users can utilise LVMs with spectral regularisation under a general model objective function $\mathcal{L}_{model}$. This makes it easier to implement LVMs with spectral regularisation and to apply the LVMs on single channel time-series applications. Given some time-series signal $x[n], n \in \mathbb{Z}^+$, the signal pre-processing, spectral regularisation, and model estimation steps are handled by the framework, thereby automating the involved aspects of the LVM task.

\subsection{Software architecture}
The \texttt{spectrally-regularised-LVMs} module is composed of five sub-modules which were designed to facilitate the steps detailed in Figure \ref{fig:package_description}. To facilitate step one from Figure \ref{fig:package_description}, the \texttt{cost\_functions} sub-module was implemented. This provides four cost function classes to the user of which two are dedicated to two simple LVM formulations, one which allows users to explicitly encode their cost function and its associated derivatives, and one which allows users to define their cost function symbolically using the SymPy package~\cite{Sympy2017} and internally determines the associated gradient and Hessian of the objective function. Additionally, a finite difference approximation scheme can be used to approximate the gradient and Hessian. The \texttt{negen\_approx} sub-module supports step one by providing estimator classes which provide access to a set of approximation functions for negentropy~\cite{Hyvarinen1998}. The \texttt{spectral\_regulariser} sub-module supports step two from Figure \ref{fig:package_description} by implementing a spectral regulariser class that can be used to determine the regularisation term given in Equation \eqref{eq:constraint} alongside its gradient and Hessian. The \texttt{helper\_methods} sub-module enables step three from Figure \ref{fig:package_description}, by providing a set of classes which assist with data pre-processing, data batch sampling, quasi-Newton update strategies~\cite{Snyman2018}, Gram-Schmidt orthogonalisation~\cite{Burden2016}, and time-series signal hankelisation~\cite{Balshaw2023}. The \texttt{spectrally\_regularised\_model} sub-module captures steps three and four from Figure \ref{fig:package_description} and provides the linear LVM model which allows users to define their training parameters and estimate the model parameters using a .\texttt{fit}$(\cdot)$ method call to an instance of the \texttt{LinearModel} class~\cite{ScikitAPI2013}. Finally, to apply the LVM as detailed in step five from Figure \ref{fig:package_description}, methods from the \texttt{LinearModel} class may be used to facilitate the transition to and from the latent space, as detailed in Figure \ref{fig:LVM_example}. A .\texttt{transform}$(\cdot)$ method call computes the transformation from the data space to the latent space, and a .\texttt{inverse\_transform}$(\cdot)$ method call performs the transformation from the latent space back to the data space. These method calls are available to instances of the \texttt{LinearModel} class. This provides a complete framework that automatically performs the necessary steps to estimate the parameters of linear LVMs with the proposed spectral regularisation term and apply the LVM encoding and decoding steps.

\subsection{Software functionalities}
\begin{figure*}[hbt!]
	\centering
	\begin{subfigure}[b]{0.45\textwidth}
		\centering
		\includegraphics[width=\textwidth]{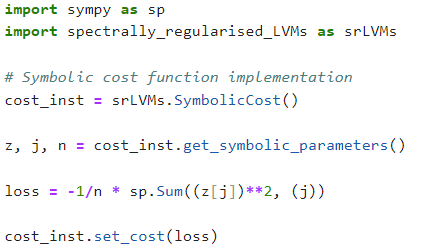}
		\caption{Symbolic objective function.}
		\label{subfig:pca-symbolic}
	\end{subfigure}
	~
	\begin{subfigure}[b]{0.45\textwidth}
		\centering
		\includegraphics[width=\textwidth]{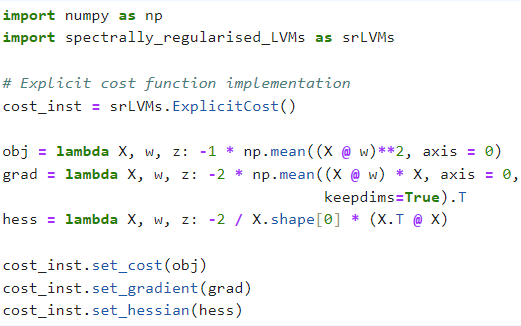}
		\caption{Explicit objective function.}
		\label{subfig:pca-explicit}
	\end{subfigure}	
		\caption{The code listings detailing how the PCA objective function given in Equation \eqref{eq:pca-objective} can be implemented symbolically or explicitly using functions that use NumPy objects~\cite{Harris2020}.}
		\label{fig:pca-objectives}
\end{figure*}

The primary functionality of the \texttt{spectrally-regularised-LVMs} package is to provide a complete framework that estimates the linear LVM model parameters that are regularised via the spectral orthogonality term. This is achieved through a cost function initialisation stage and a parameter estimation stage. Each stage is discussed in turn and code listings are given. The cost function initialisation is generalised by offering a symbolic or explicit representation to cater to a wide variety of objective functions. For this example, the PCA objective function is used
\begin{equation}\label{eq:pca-objective}
	\begin{aligned}[b]
	\mathcal{L}_{model}(\mathbf{w}_i) &= - \mathbb{E}_{\mathbf{x} \sim p(\mathbf{x})} \{ \left(\mathbf{w}_i^T \mathbf{x} \right)^2 \} \\
	&= - \frac{1}{N} \sum_{j=1}^{N} z_j^2,
	\end{aligned}
\end{equation}
where it is assumed that $\mathbf{x}$ is zero-mean and $z_j = \mathbf{w}_i^T \mathbf{x}_j$. In a symbolic representation, the indices of $z_j$ are represented for $N$ samples from $p(\mathbf{x})$. the PCA objective function can be implemented following the example in Figure \ref{subfig:pca-symbolic}. In an explicit representation, the PCA objective function can be implemented using functions that use NumPy objects as detailed in Figure \ref{subfig:pca-explicit}.

\begin{figure}[h]
	\centering
	\includegraphics[scale=0.7]{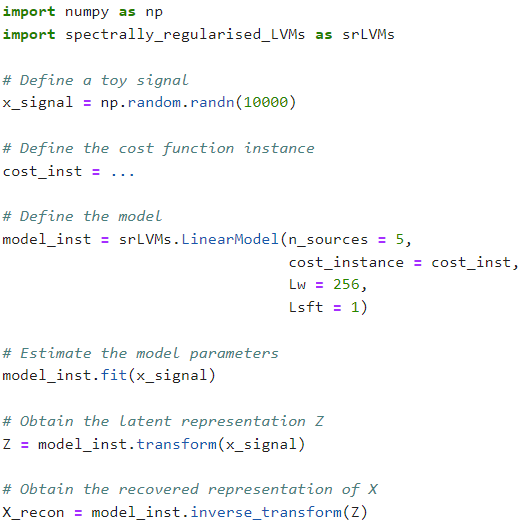}
	\caption{The code listing detailing how to perform parameter estimation in the spectrally regularised framework.}
	\label{fig:parameter-estimation}
\end{figure}

The parameter estimation stage provides alternative functionalities and advanced options to the user regarding the underlying pre-processing and model estimation steps performed. The main component is to create an instance of the \texttt{LinearModel} class and using the \texttt{.fit($\cdot$)} method to estimate the model parameters. Further information related to the arguments for the parameter estimation stage is given in the \href{https://spectrally-regularised-lvms.readthedocs.io/en/latest/}{\textcolor{blue}{documentation}}. A code listing example of this stage is given in Figure \ref{fig:parameter-estimation}.

\subsection{Software dependencies}
The \texttt{spectrally-regularised-LVMs} package is free software distributed under the MIT license, and it is open to contributions on GitHub. The package is hosted on PyPi~\cite{PyPi} to enable ease of installation, broader accessibility to the scientific community, and to avoid dependency issues when using the package in a user's local Python environment. The package is dependent on the following Python packages: NumPy~\cite{Harris2020}, Matplotlib~\cite{Hunter2007}, scikit-learn~\cite{scikit2011}, tqdm~\cite{tqdm2023} and SymPy~\cite{Sympy2017}.


\section{Illustrative example}
 To demonstrate the capabilities and detail the main functionalities of the \texttt{spectrally-regularised-LVMs} package, an illustrative example is used. In this example, steps one through five from Figure \ref{fig:package_description} are applied to the negentropy-based LVM objective function~\cite{Hyvarinen2001a}. This objective function is given as
 \begin{equation}\label{eq:negentropy-model-objective}
 	\mathcal{L}_{model}(\mathbf{w}_i) = -\left( \mathbb{E}_{\mathbf{x} \sim p(\mathbf{x})} \{ G(\mathbf{w}_i^T \mathbf{x}) \} - \mathbb{E}_{\nu \sim p(\nu)}\{ G(\nu) \} \right)^2,
 \end{equation}
where $G(\cdot)$ is a non-quadratic function and $\nu$ represents a Gaussian variable that is zero mean and unit variance~\cite{Hyvarinen1998}. This objective function is chosen as non-Gaussianity is a well-known measure for highlighting interesting information in data from rotating machinery~\cite{Antoni2006a}.

\begin{figure}[htb!]
	\centering
	\includegraphics[width=0.5\textwidth]{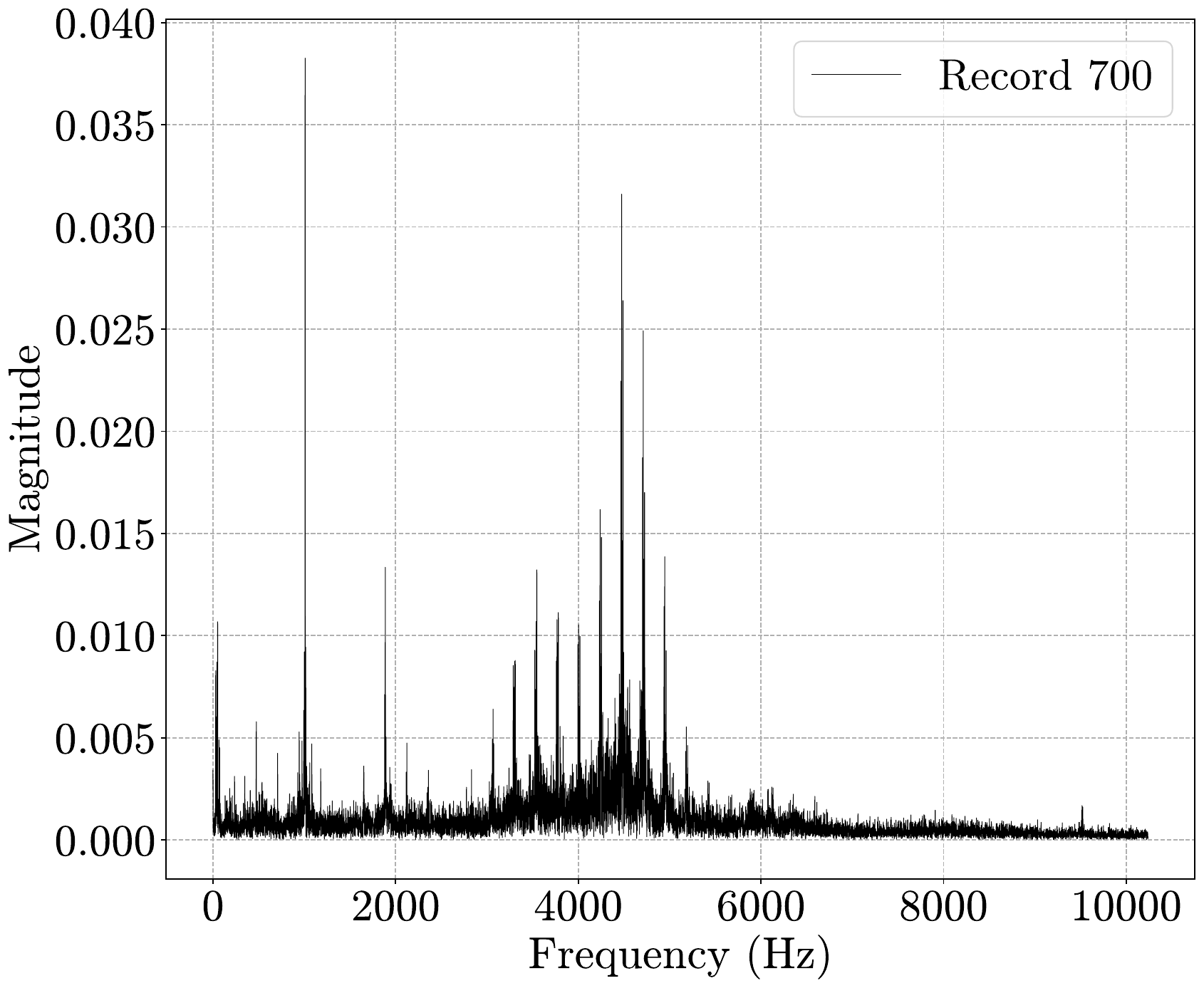}
	\caption{The signal spectrum of record 700 from the second IMS dataset.}
	\label{fig:IMS_signal}
\end{figure}

\begin{figure*}[htb!]
	\centering
	\includegraphics[scale=0.3]{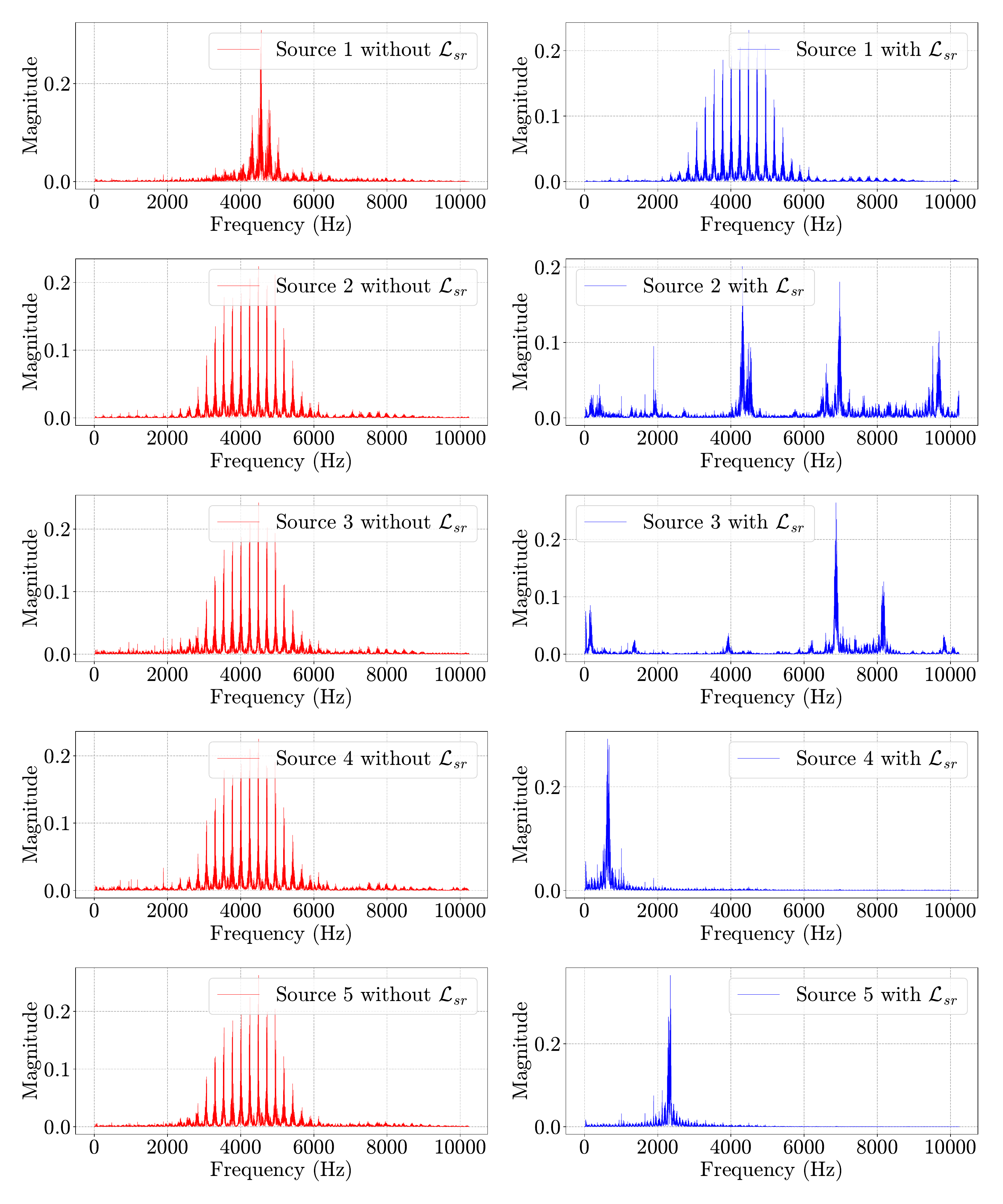}
	\caption{The spectra of the latent source signals of record 700 from the second IMS dataset without and with the addition of the proposed spectral regularisation term.}
	\label{fig:IMS_sources}
\end{figure*}

A time-series signal from the Intelligent Maintenance Systems (IMS) bearing dataset is used in this example~\cite{IMS2007}. The intention is to provide evidence of the applicability of the spectrally regularised framework on real-world data. Record 700 from the second IMS dataset is used from the first bearing channel is used as it is expected to contain a vibration signature with a faulty component~\cite{Gousseau2018}. The `Examples' folder located in the \href{https://github.com/RyanBalshaw/spectrally-regularised-LVMs/tree/main/Examples}{\textcolor{blue}{Github repository}} contains a notebook which details this step-by-step use of the proposed package for this example and contains all code for result reproducibility. In Figure \ref{fig:IMS_signal} the spectra of the record 700 is shown. The sources from the estimated linear LVM were analysed and their respective spectra with and without spectral regularisation are presented in Figure \ref{fig:IMS_sources}. Evidently, without spectral regularisation there is clear source duplication and this issue is addressed with the addition of the proposed spectral regularisation term.

\section{Impact}
The \texttt{spectrally-regularised-LVMs} package provides a framework for linear LVMs which utilises spectral regularisation for single channel time-series data applications. This was done to streamline the application process by readily adapting the parameter estimation framework to depend on a generic objective function class instance, where the user can use different model objective functions depending on their required application. Furthermore, explicit control over the model parameter estimation process is provided to ensure that different pre-processing and parameter estimation approaches can be used. This allows for efficient iteration for a given application through different optimisation strategies, e.g. steepest gradient descent, stochastic gradient descent, or constrained Newton/quasi-Newton methods~\cite{Snyman2018}. 

The Python package aims to provide researchers with a framework that not only provides a general parameter estimation methodology for linear LVMs, but also removes latent redundancy via the proposed spectral regularisation term. The proposed regularisation term is objective function agnostic in its formulation, and providing access to a Python package for LVM parameter estimation under this term is expected to be beneficial to the broader community. This will allow others to utilise the proposed methodology for various linear LVM objective formulations without having to re-implement all aspects of linear LVMs from scratch.

Finally, while many popular linear LVM applications are available in Python, e.g., PCA and ICA in scikit-learn~\cite{scikit2011}, these implementations cater to a general class of LVM applications, and are not focused on the application of the methodologies to single channel time-series data. The \texttt{spectrally-regularised-LVMs} package provides direct access to the linear LVM framework and caters to the referenced methodology application, which is expected to offer fruitful avenues for future research.

\section{Conclusion}
In this paper, a modular linear LVM framework is proposed and implemented in Python. This framework offers customizable objective functions, facilitates the estimation of the model parameters, and  uses a novel spectral regularisation term to combat source duplication. By making the \texttt{spectrally-regularised-LVM} framework accessible to the broader community it makes the application and development of new methods easier and we firmly believe it will be beneficial for future research into LVMs for single channel time-series data. Finally, it is envisaged that future research pursuits into single channel time-series data applications will be stimulated and benefit from the \texttt{spectrally-regularised-LVMs} package. 

\section*{Acknowledgements}
The authors gratefully acknowledge the support that was received from AngloGold Ashanti in the execution of this research.



\bibliographystyle{elsarticle-num} 
\bibliography{Bibliography}


%
%
%

%
%

\clearpage

\section*{Required Metadata}
\label{}

\section*{Current code version}
\label{}

\begin{table}[h!]
	\caption{Code metadata}
	\begin{tabular}{|l|p{6.5cm}|p{6.5cm}|}
		\hline
		\textbf{Nr.} & \textbf{Code metadata description} & \textbf{Please fill in this column} \\
		\hline
		C1 & Current code version & v0.1.3 \\
		\hline
		C2 & Permanent link to code/repository used for this code version & \url{https://github.com/RyanBalshaw/spectrally-regularised-LVMs} \\
		\hline
		C3  & Permanent link to Reproducible Capsule & N/A \\
		\hline
		C4 & Legal Code License & MIT license (MIT) \\
		\hline
		C5 & Code versioning system used & git \\
		\hline
		C6 & Software code languages, tools, and services used & Python3 \\
		\hline
		C7 & Compilation requirements, operating environments & $\text{Python }\geq 3.10$, 
		$\text{Numpy }\geq 1.23.1$, $\text{Matplotlib }\geq 3.5.2$, $\text{SciPy }\geq 1.8.1$, 
		$\text{scikit-learn }\geq 1.1.2$, $\text{tqdm }\geq 4.64.1$, and $\text{SymPy }\geq 1.1.1$ \\
		\hline
		C8 & If available Link to developer documentation/manual & See: \url{https://spectrally-regularised-lvms.readthedocs.io/en/latest/} \\
		\hline
		C9 & Support email for questions & ryanbalshaw1@gmail.com \\
		\hline
	\end{tabular}
	\label{tab:metadata} 
\end{table}

%
%

\end{document}